\documentclass{article}
\usepackage{spconf,amsmath,graphicx}

\usepackage{enumitem}
\usepackage{amssymb,amsfonts}
\usepackage{algorithmic}
\usepackage{graphicx}
\usepackage{textcomp}
\usepackage{amsthm}
\usepackage{amscd}
\usepackage{siunitx}
\usepackage{graphicx}
\usepackage{booktabs}
\usepackage[ruled,linesnumbered]{algorithm2e}
\usepackage{bm}
\usepackage{dsfont}
\usepackage{multirow}
\usepackage{nicefrac}
\usepackage{rotating}
\usepackage{makecell}
\usepackage{colortbl}
\usepackage{subcaption}
\usepackage{url}
\usepackage[numbers,sort,compress]{natbib}

\setlist{nosep, leftmargin=14pt}

\newcommand{\ours}{THC\xspace}
\newcommand{\clustering}{BCluster\xspace}

\title{\mbox{Transformer-Based Hierarchical Clustering for Brain Network Analysis}}
\name{{Wei Dai$^{\dagger}$ \quad  Hejie Cui$^{\ddagger}$ \quad Xuan Kan$^{\ddagger}$ \quad  Ying Guo$^{\ddagger}$\quad  Sanne van Rooij$^{\ddagger}$\quad  Carl Yang$^{\ddagger *}$\thanks{*Corresponding author: Carl Yang $<$j.carlyang@emory.edu$>$}}}
\address{$^{\dagger}$ Stanford University, 
    $^{\ddagger}$ Emory University}
    
\begin{document}
\maketitle
\begin{abstract}
Brain networks, graphical models such as those constructed from MRI, have been widely used in pathological prediction and analysis of brain functions. Within the complex brain system, differences in neuronal connection strengths parcellate the brain into various functional modules (network communities), which are critical for brain analysis. However, identifying such communities within the brain has been a nontrivial issue due to the complexity of neuronal interactions. In this work, we propose a novel interpretable transformer-based model for joint hierarchical cluster identification and brain network classification. Extensive experimental results on real-world brain network datasets show that with the help of hierarchical clustering, the model achieves increased accuracy and reduced runtime complexity while providing plausible insight into the functional organization of brain regions. The implementation is available at \url{https://github.com/DDVD233/THC}.
\end{abstract}
\section{Introduction}

Graph is a ubiquitous form of data as it captures multiple objects and their interactions simultaneously. It is widely used for representing complex systems of related entities \cite{hu2020open, xu2019powerful}. Brain network is a special kind of graph constructed from MRI images. In brain networks, the anatomical areas named ``Region of Interests'' (ROIs) are represented as nodes, while connectivities between ROIs are represented as links. Partition atlas defines the set of ROIs in a particular brain network. In recent decades, abundant works have shown strong associations between linking imaging-based brain connectivity and demographic characteristics or mental disorders~\cite{cui2022braingb, jiang2020hi,10021060, yu2022learning,yang2022data}.

Both shallow machine learning models like M2E \cite{M2ELiu} and deep models like graph neural networks (GNNs) \cite{li2020braingnn} are researched in the area of brain network analysis. Shallow models exhibits inferior performance as compared to deep models \cite{cui2022braingb}, but GNNs also suffer from over-smoothing \cite{hamilton2020graph}, which limits their ability to model long-distance interactions. Transformers, on the other hand, have recently emerged as a promising approach for various tasks \cite{DBLP:conf/naacl/DevlinCLT19, DBLP:conf/iclr/DosovitskiyB0WZ21}, including predictions on graph data \cite{san}. Graph based transformers utilizes pairwise attention across full graphs, unlike GNNs, which only propagate node embeddings locally. BrainTransformer \cite{kan2022bnt} employs transformer on brain networks and demonstrates state-of-the-art performance for brain network analysis.

ROIs in Brain networks are inherently hierarchically clustered \cite{akiki2019determining}. In typical brain network analysis \cite{genderfunction}, clustered ROIs form communities, with each representing a particular functional module. These functional modules are then further organized into larger functional modules, with each responsible for a more general function. This arrangement creates a hierarchical ``module-in-module'' structure \cite{meunier2009hierarchical}. Functional modules provide critical information with regard to downstream tasks, and alterations in community patterns sometimes signal pathological lesions \cite{alexander2012discovery}. Therefore, learning a globally shared cluster assignment with the awareness of downstream tasks is beneficial for both model optimization and clinical examinations. However, learning such hierarchical cluster representations is difficult. Shallow methods proposed to detect brain communities are mostly based on the Louvain algorithm \cite{sporns2016modular} and Lloyd algorithm \cite{nanetti2009group}. These correlation-based methods fail to capture higher-order connectivity patterns between brain regions \cite{louvian_bad}. Some GNN models are proposed to detect communities within the brain \cite{li2020braingnn}, but these models suffer from the over-smoothing problem of GNN, limiting their ability to aggregate and identify multi-hop connectivity patterns. 

To solve the aforementioned challenges, we propose a Transformer-based Hierarchical Clustering model, abbreviated as \ours, that is tailored for brain network analysis. We highlight three main contributions of our clustering model. First, we offer an end-to-end transformer based approach to learn clustering assignments. Through pairwise attention, a clustering layer, \clustering, and a transformer encoder collaboratively learn a globally shared clustering assignment that is continuously tuned to downstream tasks. \clustering enhances the model's performance, reduces run time complexity, while also providing clinical insights. Second, we propose a hierarchical structure for the clustering model, enabling the model to learn more abstract, higher-level cluster representations by combining lower-level modules. Each clustering layer is attached to a distinct readout module, which allows the model to effectively utilize the cluster embeddings of every layer. Last but not least, we redesign the attention mechanism of the transformer with stochastic noise, which enhances its cluster learning capability. We compare our model's performance with SOTA models and perform clustering analysis with the ground truth community labels \cite{Akiki_Abdallah_2019}. Empirical analysis demonstrates the superior prediction power of our model, and the assignment produced by \ours aligns well with the ground-truth functional module labels. 
\section{Related Works}
\noindent \textbf{Deep Learning for Brain Network Analysis.}
Deep learning models have gained widespread application in brain network analysis due to their ability to discover shared patterns across samples with diverse characteristics \cite{xuan2022fbnetgen}. To this end, IBGNN \cite{cui2022interpretable} has emerged as one of the first interpretable graph neural networks (GNNs) for brain networks. Graph-based transformers such as SAN \cite{san} and Graphormer \cite{graphormer} have been recently introduced as promising alternatives to GNN. Brain Network Transformer \cite{kan2022bnt} presents a novel approach that includes unique readout functions and optimal positional embeddings and attention mechanisms for brain networks. BrainNN \cite{zhu2022joint} and CroGen \cite{luo2022multi} explore the integration of multiview brain networks. However, despite the promise of the existing models, they lack the ability to provide the clustering information that is essential for brain network analysis.

\noindent \textbf{Cluster-aware GNNs.}
Brain networks have overlapping communities \cite{overlap_community}. To cluster nodes, DiffPool \cite{diffpool} uses an auxiliary GNN, GRACE \cite{yang2020grace} utilizes stable states of repeated propagation, whereas HoscPool \cite{DBLP:conf/cikm/DuvalM22/HoscPool} uses probabilistic motif spectral clustering. GNN models like BrainGNN \cite{li2020braingnn} detect communities, but over-smoothing limits their ability to identify complex cluster structures \cite{chen2020measuring}.
\section{Transformer-based Hierarchical Clustering (THC)}

\begin{figure*}[!ht]
    \centering
    \includegraphics[width=0.9\linewidth]{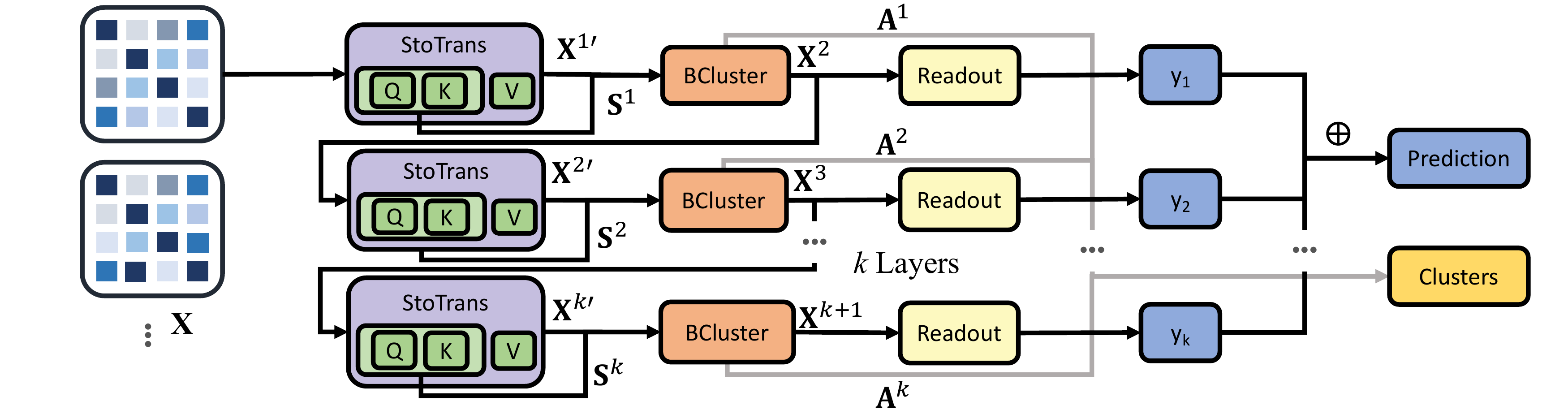}
    \caption{The overall framework of our proposed model \ours. }
    \label{fig:model}
    \vspace{-10pt}
\end{figure*}

\noindent \textbf{Problem Definition}
Our model's input is a weighted adjacency matrix $\mathbf{X} \in \mathbb{R}^{V \times V}$ of a brain network, where $V$ is the number of nodes (ROIs) as defined in the network. The objective of the model is to predict the sample class $y$ and a k-layer hierarchical cluster assignment $(\mathbf{A}^1, \cdots, \mathbf{A}^{k})$. As shown in Figure \ref{fig:model}, the overall model can be broken down into three modules: a Stochastic Transformer (StoTrans) encoder, a Brain Clustering (\clustering) module and a Readout.

\noindent \textbf{Transformer Based Clustering.} 
The stacked \clustering layer learns globally shared cluster assignments $(\mathbf{A}^1, \dots, \mathbf{A}^k)$, where $k$ is the number of model layers. Each assignment $\mathbf A^i \in (C_i, C_{i+1})$ reduces $\mathbf X^{i} \in (C_i, C_i)$ into $\mathbf X^{i+1} \in (C_{i+1}, C_{i+1})$, where $C_{i+1} \in (C_1, \cdots, C_{k})$ are the \textit{hyperparameters} denoting the cluster sizes of the $i^{th}$ layer. Specifically, at layer i, the adjacency matrix $\mathbf{X}^i$ is first fed into the transformer encoder, which learns attention matrix $\mathbf{S}^{(i, m)}$
\vspace{-5pt}
\begin{equation}
    \mathbf{S}^{(i, m)}  = \frac{\mathbf W_{\mathcal{Q}}^{i, m} \mathbf X^{i} (\mathbf W_{\mathcal{K}}^{i, m} \mathbf X^{i})^{\top}}{\sqrt{d_{\mathcal{K}}^{i, m}}}
    \label{eq:s_im}
\end{equation}
where $\mathbf W_{\mathcal{Q}}^{i, m}$ and $\mathbf W_{\mathcal{K}}^{i, m}$ are learnable parameters. A node embedding $\textbf{X}^{i'}$ is also learned through vanilla transformer propagation using the attention matrix \cite{vaswani2017attention}. 
\vspace{-3pt}
\begin{equation}
    \mathbf{X}^{i'} = \frac{1}{M}\sum_{m=1}^{M} \operatorname{Softmax}( \mathbf{S}^{(i, m)} ) \mathbf{X}^{i}  \mathbf{W}_{\mathcal{V}}^{i} 
\end{equation}
where $\mathbf{W}_{\mathcal{V}}^{i}$ is a learnable parameter. Then, a batch-wise shared clustering assignment is obtained by 
\begin{equation}
    \mathbf{A}^i = \operatorname{Softmax}\left(\frac{1}{M}\sum_{m=1}^{M} \mathbf{S}^{(i, m)} \mathbf{W}_{\mathcal{A}}^{i}\right)
\end{equation}
where $\mathbf{W}_{\mathcal{A}}^{i}$ is a learnable parameter and the $\operatorname{Softmax}$ is applied row-wise. The learned node embedding $\mathbf X^{i'}$ is fed into an encoder MLP layer, followed by the multiplication of the globally shared clustering assignment $\mathbf{A}^i$:
\vspace{-5pt}
\begin{equation}
    \mathbf{X}^{i+1} = \text{MLP}(\mathbf{X}^{i'}) \cdot \mathbf{A}^i.
\end{equation}
\vspace{-10pt}

The main objective of \clustering is maximizing the mutual information between the pre-clustered embedding $\mathbf{X}^{i'}$ and cluster embedding $\mathbf{X}^{i+1}$. This objective encourages the model to cluster similar nodes as the information loss of combining nodes with similar representations is also minimal. Inspired by \cite{cui2022interpretable}, we supervise the parameter $\mathbf{W}_{\mathcal{A}}$ jointly with the full model through cross-entropy loss. Two losses, sparsity loss and element-wise entropy loss are added to regularize the number of clusters each node is assigned to
\vspace{-5pt}
\begin{align}
    \mathcal{L}_\textsc{sps}&=\sum\nolimits_{i, j} \bm A_{i, j}, \\
    \mathcal{L}_\textsc{ent} &= -(\bm S \log (\bm A)+(1-\bm A) \log (1-\bm A)).
\end{align}
\vspace{-15pt}

The final objective is the sum of all three losses. The \clustering layer has the following favorable properties: 
\begin{enumerate}
    \item \textit{The Assignment is Globally Shared.} During the training process, the assignment is shared batch-wise and optimized jointly with the model. After training is complete, a final globally shared assignment is obtained by averaging across the assignments of all samples. 
    \item \textit{The Assignment is Differentiable.} Each node/cluster $\mathcal{V}^{i,j} \in \mathbf{X}^i$ at layer $i$ is assigned a vector of probabilities $(p_{\mathcal{V}^{i,j}, 1}, \cdots,$ $p_{\mathcal{V}{i,j}, k})$, with $p_{\mathcal{V}{i,j}, k}$ denoting the probability of assigning $\mathcal{V}^{i,j}$ to $k^{th}$ cluster. This soft assignment is differentiable, allowing the model to optimize the assignment directly through gradient descent. 
    \item \textit{Clustering Reduces Computational Complexity.} At $i^{th}$ layer, the computational complexity of each transformer encoder is $O(n^2 d + nd^2)$, where $n$ represents the number of nodes and d represents the number of features. The clustering reduces the complexity to $O(k^2 d + kd^2)$, where k is the cluster count and $k << n$. 
    \item \textit{The Assignment is Squashable. } A hierarchical clustering assignment produced by \clustering can be flattened into one-layer clustering with a linear number of matrix multiplications. The k-layer hierarchical assignments $(\mathbf{A}^1, \dots, \mathbf{A}^k)$ are reduced into one assignment $\mathbf{A}^{flat}$ by multiplication, $\mathbf{A}^{flat} = \prod_{i=1}^k \mathbf{A}^i \label{eq:flatten}$. 
\end{enumerate}

\noindent \textbf{Hierarchical Structure.}
The model further groups the cluster embeddings from the previous layer into new, larger clusters, generating a hierarchical structure. A readout layer is attached to each \clustering layer. At layer $i$, an MLP Readout is attached to the \clustering layer of that layer $\mathbf{X^{i}}$ and outputs a prediction $y_i$. This allows the model to effectively ensemble cluster embeddings from different layers. The final prediction $y$ is the average of all outputs of each Readout layer, where $y = \frac{1}{k} \sum_{i=1}^k y_i$.

\noindent \textbf{Stochastic Noise.} 
To allow \clustering to effectively learn the clustering assignment without falling into the trivial situation of replicating the results from the attention matrix, we redesign attention matrix $\mathbf{S}^{(i, m)}$ with stochastic noise
\vspace{-3pt}
\begin{equation}
    \mathbf{S}^{(i, m)'}  = \mathbf{S}^{(i, m)} + log(\frac{\mathbf{B}^{i, m}}{\mathbf{I} - \mathbf{B}^{i, m}})
    \label{eq:s_im2}
\end{equation}
where $\mathbf{B} \sim \mathbf{S}^{i, m}$ is a random variable chosen uniformly from range (0, 1) and $\mathbf{I} \sim \mathbf{S}^{i, m}$ is a diagonal matrix of ones. 
\section{Experiments}

\noindent \textbf{Datasets. }
The method is tested on two fMRI datasets. 

\begin{enumerate}
    \item \textit{Adolescent Brain Cognitive Development Study (ABCD)}: This research enrolls children aged 9 to 10 years old. Repeated MRI scans are used to track each child until early adulthood \cite{ABCD}. Each sample is labeled with its biological sex. 7,901 children are involved in the study, with 3,961 (50.1\%) of them being female. The parcellation is based on HCP 360 atlas \cite{hcp360}.
    \item \textit{Autism Brain Imaging Data Exchange (ABIDE)}: This dataset collects resting-state functional magnetic resonance imaging (rs-fMRI) \cite{abide}. It contains brain networks from 1009 subjects, with 516 (51.14\%) being labeled as Autism spectrum disorder (ASD) patients (positives). The region is defined based on Craddock 200 atlas \cite{craddock2012whole}.
\end{enumerate} 

\begin{table}[th]
\centering
\small
\caption{Performance comparison of our model with different baselines in terms of AUROC and accuracy.}
\label{tab:performance}
\resizebox{1.0\linewidth}{!}{
\begin{tabular}{ccccccccc}
\toprule
\multirow{2.5}{*}{Type} & \multirow{2.5}{*}{Method} &\multicolumn{2}{c}{\bf Dataset: ABIDE} & \multicolumn{2}{c}{\bf Dataset: ABCD}\\
\cmidrule(lr){3-4} \cmidrule(lr){5-6} 
& & {AUROC} & {Accuracy} &  {AUROC} & {Accuracy} \\
\midrule
\multirow{3}{*}{\thead{Graph\\Transformer}}
& SAN & 71.3±2.1 & 65.3±2.9 & 90.1±1.2 & 81.0±1.3 \\\
& VanillaTF & 76.4±1.2 & 65.2±1.2&94.3±0.7 & 85.9±1.4  \\
& Graphormer & 63.5±3.7 & 60.8±2.7 &  89.0±1.4 & 80.2±1.3  \\
\midrule
\multirow{3}{*}{\thead{Fixed\\Network}}
&BrainGNN             & 62.4±3.5 & 59.4±2.3 &  OOM      & OOM \\
& BrainGB             & 69.7±3.3 & 63.6±1.9 &  91.9±0.3 & 83.1±0.5 \\
& BrainNetCNN         & 74.9±2.4 & 67.8±2.7 &  93.5±0.3 & 85.7±0.8  \\
\midrule
\multirow{3}{*}{\thead{Pooling \& \\ Clustering}} 
& DiffPool            & 68.3±0.9 & 61.3±1.7 & 83.2±1.2 & 75.5±1.4 \\
& \ours+ Lin. Cluster & 71.8±1.1 & 64.8±1.3 & 91.7±0.5 & 82.6±1.8 \\
& \ours+ No Cluster   & 75.6±1.2 & 68.0±1.4 & 94.5±0.7 & 86.4±1.1 \\
\midrule
\multirow{1}{*}{Ours} 
& \ours Full & \textbf{79.76±1.1} & \textbf{70.6±2.2} & \textbf{96.2±0.5} & \textbf{89.4±0.4} \\
\bottomrule
\end{tabular}
}
\vspace{-15pt}
\end{table} 

\begin{figure*}[!t]
\centering
\small
\centering
\begin{minipage}{\textwidth}
    \centering
    \begin{subfigure}{0.246\textwidth}
        \includegraphics[width=\textwidth]{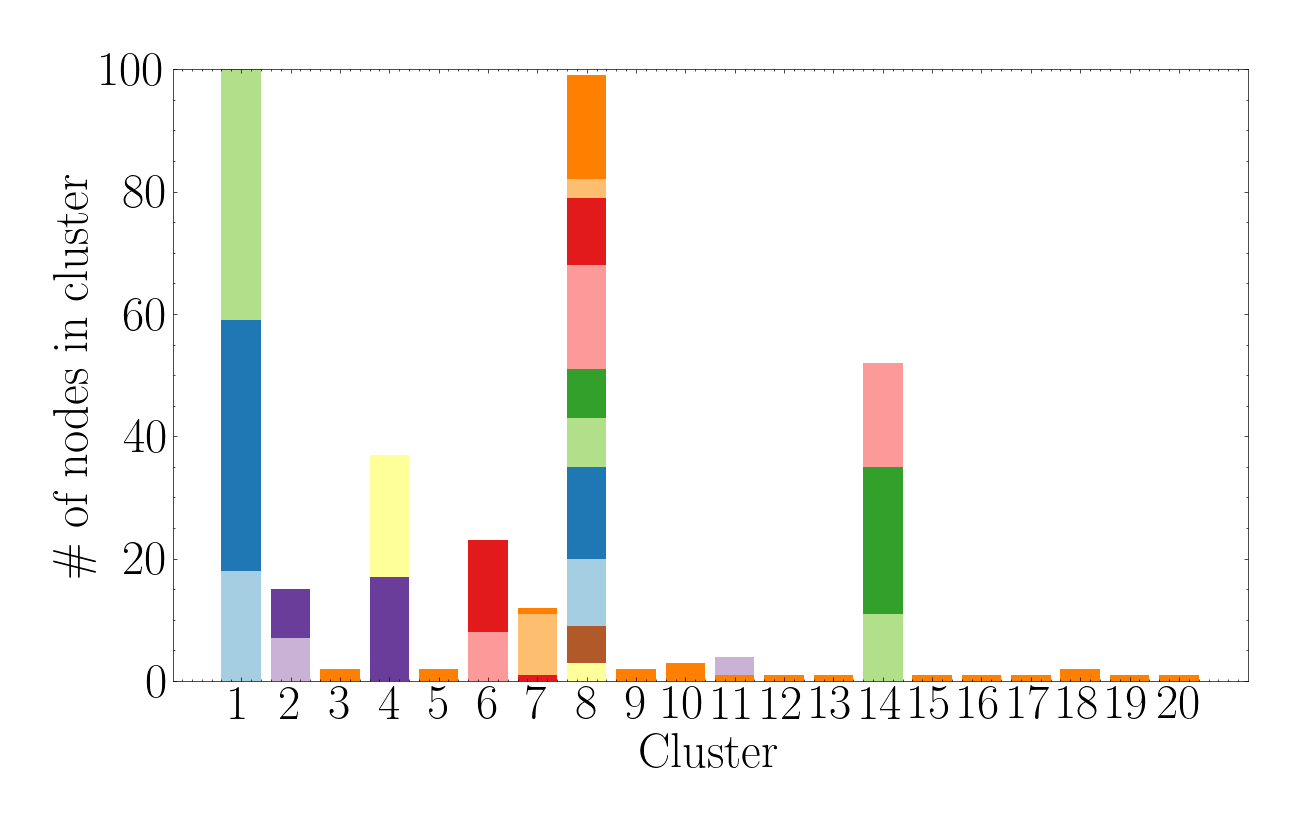}
        \caption{Louvain, 20 clusters}
    \end{subfigure}
    \begin{subfigure}{0.246\textwidth}
        \includegraphics[width=\textwidth]{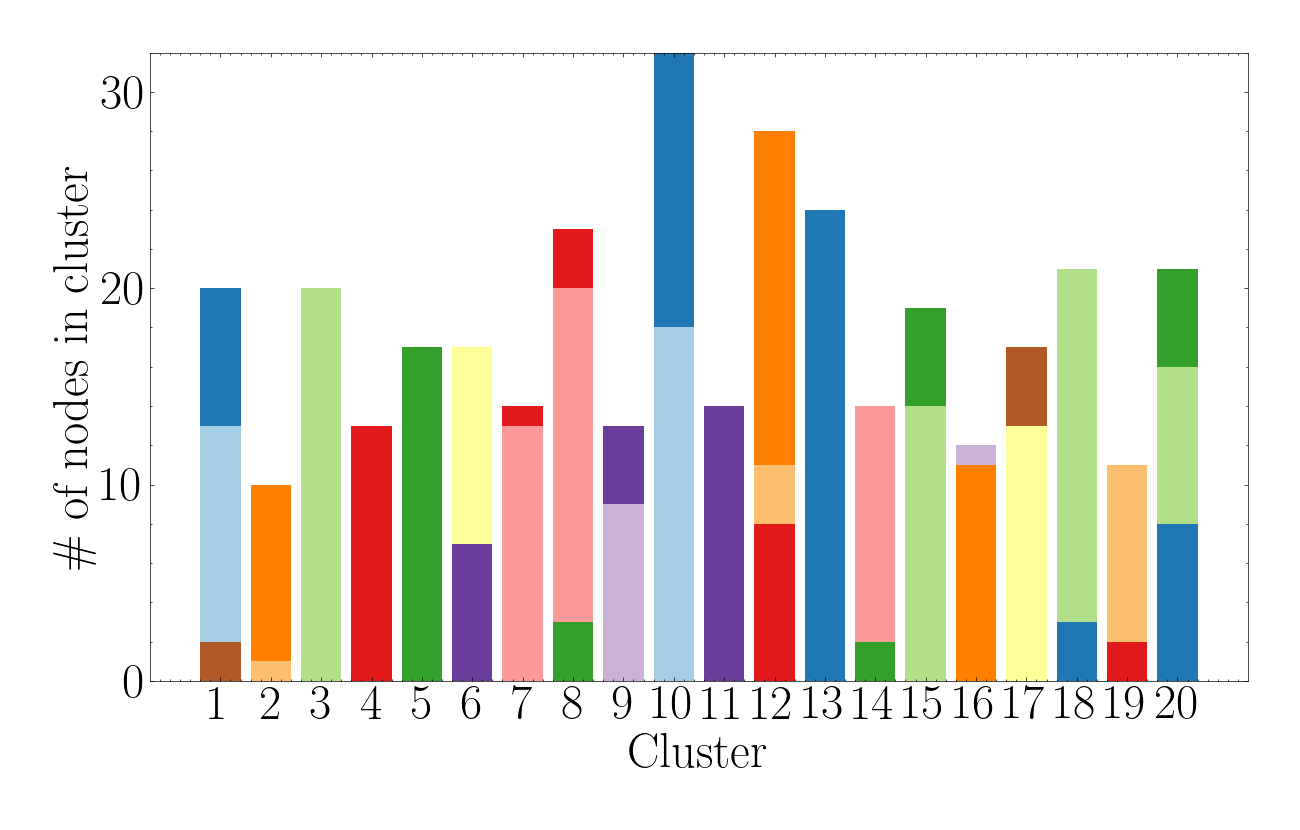}
        \caption{Ours, 20 clusters}
    \end{subfigure}
    \begin{subfigure}{0.246\textwidth}
        \includegraphics[width=\textwidth]{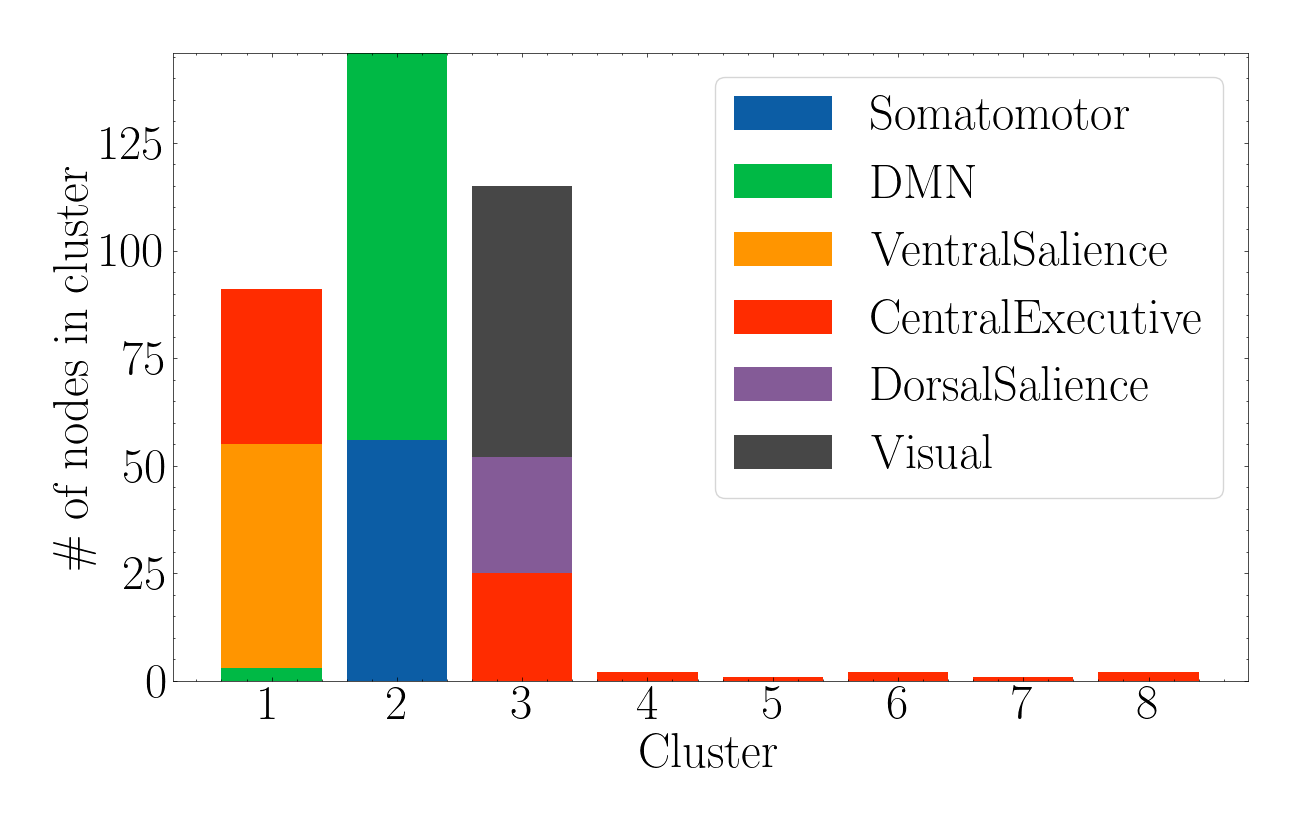}
        \caption{Louvain, 8 clusters}
    \end{subfigure}
    \begin{subfigure}{0.246\textwidth}
        \includegraphics[width=\textwidth]{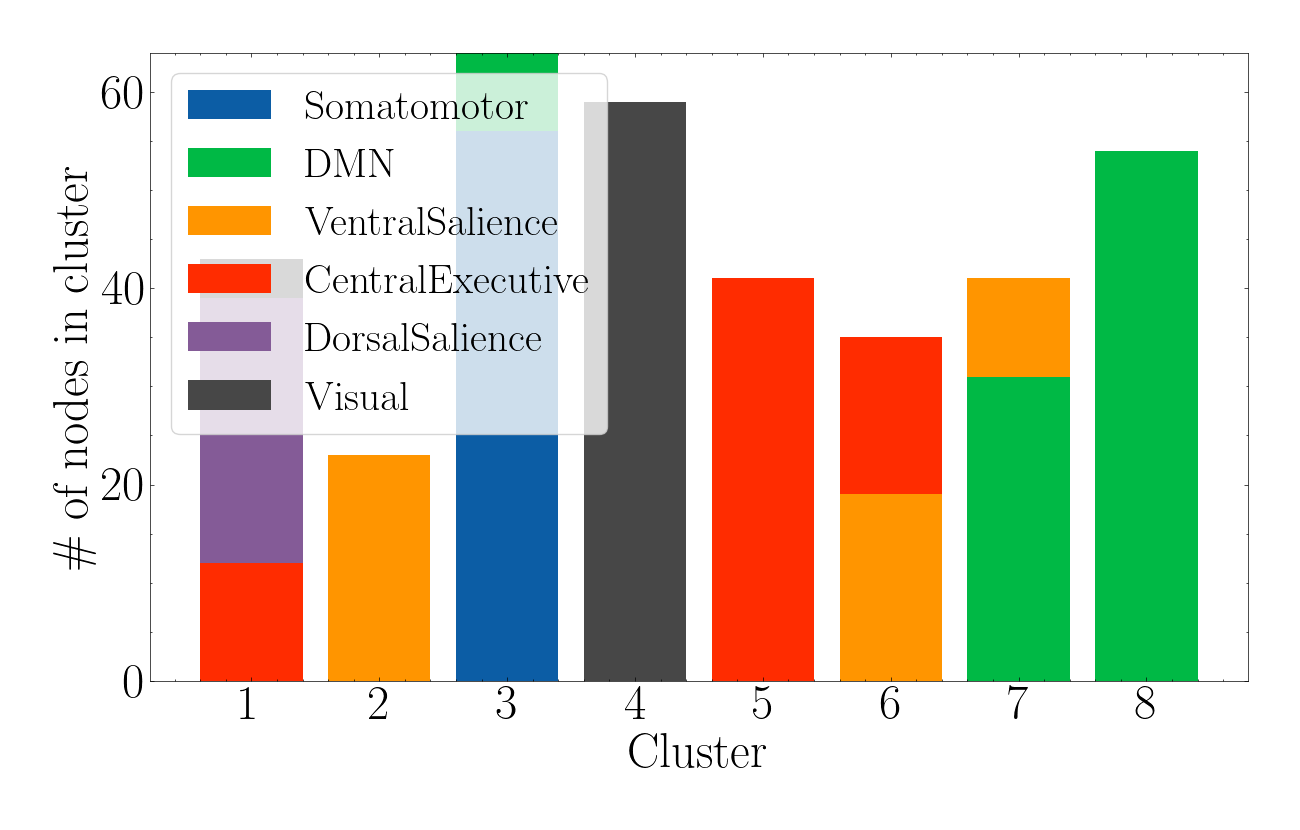}
        \caption{Ours, 8 clusters}
    \end{subfigure}
\end{minipage}
\caption{Hierarchical 2-layer cluster assignments with 20 and 8 clusters produced by \ours and Louvain. In each cluster, nodes are colored in accordance with their ground-truth cluster labels. The legend for 20-cluster plots are omitted for simplicity. }
\label{fig:cluster_community}
\vspace{-10pt}
\end{figure*}

\noindent \textbf{Baselines and Experiment Setup. }
We compare the \ours model with two kinds of baselines, as demonstrated in Table \ref{tab:performance}. Specifically, SAN \cite{san}, Graphormer \cite{graphormer} and Vanilla Transformer are transformer-based baselines, with the first two designed for graph prediction, and the vanilla transformer is a two-layer transformer based on the original implementation \cite{vaswani2017attention}. Three methods optimized for brain networks, BrainGNN \cite{li2020braingnn}, BrainGB \cite{cui2022braingb}, and BrainNetCNN \cite{kawahara2017brainnetcnn}, are also included. Our model is experimented with two layers and clustering sizes 90, 4 and 20, 4 for ABCD and ABIDE dataset respectively.
We split the dataset randomly into a training set, a test set and a validation set on a 7:2:1 ratio. AUROC and accuracy are used over a 5-fold run as metrics. For all methods, the test performance on the epoch with the highest average AUROC on the validation set is recorded. 

\noindent \textbf{Performance Analysis. }
Performance results are recorded in Table \ref{tab:performance}. Compared with the baseline methods, it is shown that our model achieves the best performance on both datasets, with up to 3.2\% and 1.7\% higher AUROC on ABIDE and ABCD datasets over all transformer and GNN-powered models. Out of the 7 baselines, BrainGNN is the only model that generates community clustering. Nevertheless, its need of memory prevents its application on large graphs like ABCD. 

To understand how our clustering method affects the model performance, we designed two ablative variants of our model: one with the clustering layer removed (\ours+ No Cluster), and one with \clustering swapped with a simple linear layer (\ours+ Lin. Cluster). We further compare the performance of our model with a popular pooling model, DiffPool \cite{diffpool}. The results are shown in the Pooling \& Clustering section of Table \ref{tab:performance}. Compared with the no clustering model, we observe that our full model gives a significant performance boost of up to 1.7\% in AUROC and 3.0\% in accuracy. Furthermore, we discover that a naive linear clustering negatively affects performance, proving that clustering assignments cannot be modeled in a simple way.

\begin{table}[h]
\caption{Runtime comparison of our model with different baselines (in minutes) averaged over 5 runs.}
\resizebox{1.0\linewidth}{!}{
\begin{tabular}{ccccccccc}
\toprule
Dataset & SAN & Graphormer & Vanilla TF & \textbf{\ours} \\
\midrule
ABCD & 908.05±3.6 & 4089.86±5.7 & 36.26±2.12 & \textbf{27.31±0.47} \\
ABIDE & 93.01±0.96 & 133.52±0.54 & 2.32±0.10 & \textbf{1.81±0.03} \\
\bottomrule
\end{tabular}
\label{runtime}
}
\end{table}

\noindent \textbf{Runtime Analysis.} We compare the run time of our model with transformer baselines on an RTX 8000 GPU and record the results in Table \ref{runtime}. Our model runs faster than all baselines with a margin of at least 24.7\% on ABCD and 22.0\% on ABIDE, which demonstrate that our model can effectively reduce runtime complexity through the efficient aggregation of node embeddings into cluster embeddings of smaller sizes. 
\section{Clustering Analysis}

To investigate the quality of the model's clustering assignment, we run our model with 2 layers of cluster size 20 and 8, and create a bar plot of each cluster, as shown in Figure \ref{fig:cluster_community}. The nodes within the 20 clusters assignment and 8 clusters assignments, represented by blocks in the cluster bar, are colored with respect to the AAc architecture at 20 modules and 6 modules definition by \cite{Akiki_Abdallah_2019} respectively. In these bar plots, cluster purity is represented by the percentage of the majority color of that cluster. The results reveal that the proposed clustering method is able to produce clustering assignments similar to the existent labels for all functional modules. Among all the plotted clusters, cluster 2, 3, 4, 6 and 8 have purities of over 90 percent. Similar patterns are observed in the 20 clusters plot. Misclassified samples also provide interesting insights. In the 8 clusters sample, some Ventral Salience Network nodes are misclassified as Default Mode Network (DMN). This is supported by studies showing some areas of these two communities share similar functions \cite{doll2015mindfulness}.

\begin{table}[ht]
\centering
\small
\caption{Cluster quality comparison of our model with baselines. Clustering results are compared with the labels by \cite{Akiki_Abdallah_2019}.}
\label{tab:clustering}
\begin{tabular}{ccccccccc}
\toprule
Method & Purity & NMI & Homogeneity \\
\midrule
Lloyd & 0.842±0.003 & 0.738±0.001 & 0.711±0.002 \\\
Louvain & 0.671±0.099 & 0.647±0.052 & 0.648±0.087 \\
GRACE & 0.717±0.067 & 0.685±0.022 & 0.722±0.053 \\
\midrule
\ours & \textbf{0.889±0.008} & \textbf{0.742±0.004} & \textbf{0.783±0.017} \\
\bottomrule
\end{tabular}
\end{table} 

To quantitatively analyze the result, we further compare our method with three clustering methods, Lloyd's algorithm \cite{Lloyd}, Louvain method \cite{louvain}, and GRACE \cite{yang2020grace}, in terms of purity, normalized mutual information (NMI) and homogeneity, as shown in Table \ref{tab:clustering}. Specifically, for a cluster assignment $C$ and a ground truth definition $T$ partitioning $N$ nodes, purity $p_{\text{purity}}$ is defined as the percentage of the majority truth cluster within $C$: $p_{\text{purity}} = \sum\limits_{c \in C} \frac{1}{N} \max\limits_{t \in T}  | c~\cap~t|$. The NMI score $p_{\text{nmi}}$ and homogeneity score $p_{\text{homo}}$ are given as $p_{\text{nmi}} = H(C) - H(C|T) - E(H(C) - H(C|T))$ and $p_{\text{homo}} = 1 - \frac{H(C|T)}{H(C)}$ respectively, where $H$ denotes the entropy and $E$ denotes expected value. We can observe that the proposed THC produces clusters with the highest quality among both deep and shallow baselines. The purity of THC averaged over 5 k-fold runs is 88.9\%, which is 4.7\% higher than the highest baseline. Similarly, the homogeneity score of THC surpasses the highest baseline by a margin of 7.2\%. These results further support that the proposed clustering THC can capture the complex structure of brain networks. 
\section{Conclusion}

In this work, we present \ours, an interpretable transformer model for brain network analysis. Hierarchical clustering layers are designed based on the attention mechanism of the transformer encoder to learn a global clustering assignment in an end-to-end manner. A future direction of our work is to investigate the effect of overlapping communities on the clustering result as well as the lesion predictions. 

\section{Compliance with Ethical Standards}
The ABCD and ABIDE datasets employed in this study are owned by a third-party organization, where informed consent was obtained for all subjects. The data processed is anonymous with no personally identifiable information. All studies are conducted according to the Good Clinical Practice guidelines and U.S. 21 CFR Part 50 (Protection of Human Subjects), and under the approval of Institutional Review Boards. 

\bibliographystyle{IEEEtranN}
\bibliography{base}

\end{document}